\pdfoutput=1

\documentclass[11pt]{article}

\usepackage{ACL2023}

\usepackage{times}
\usepackage{latexsym}

\usepackage[T1]{fontenc}

\usepackage[utf8]{inputenc}

\usepackage{microtype}

\usepackage{inconsolata}

\usepackage{todonotes}
\usepackage[capitalize]{cleveref}

\def\paperDraft{}
\ifdefined\paperDraft%
    \def\ad#1{{\color{teal}[Amandine: \textit{#1}]}}
    \def\ma#1{{\color{magenta}[Maxime]: \textit{#1}}}
    \def\eb#1{{\color{orange}[Ellen: \textit{#1}]}}

\else
    \def\ad#1{}
    
    \def\ma#1{}
    
    \def\eb#1{}
    
\fi

\title{``Wait, did you mean the doctor?’’: Collecting a Dialogue Corpus for Topical Analysis}

\author{Amandine Decker$^{1,2}$, Vincent Tourneur$^{1}$, Maxime Amblard$^{1}$ \and Ellen Breitholtz$^2$\\
        $^1$Université de Lorraine, CNRS, Inria, LORIA, F-54000 Nancy, France \\ \texttt{\{amandine.decker, vincent.tourneur, maxime.amblard\}@loria.fr} \\ $^2$University of Gothenburg, CLASP \\ \texttt{ellen.breitholtz@ling.gu.se}}

\begin{document}
\maketitle

\section{Introduction}

Dialogue is at the core of human behaviour and being able to identify the topic at hand is crucial to take part in conversation. 
Nevertheless, from a scientific point of view, the notion of topic is somewhat elusive. \citet{mittwoch2002clause} and \citet{geoffrey2004so} focus on topic shift markers, while \citet{howe1991topic} introduces \textit{topic transition relevance places}, inspired from \citet{SimplestSystematics}. \citet{hsueh-etal-2006-automatic} and \citet{georgescul-etal-2008-comparative} propose topic segmentation methods for meetings, \textit{i.e.}, rather organised exchanges. Yet, there are few accounts of the topical organisation in casual dialogue and of how people recognise the current topic.


In a conversation, topics can follow each other in a linear way or gradually drift to another. Interesting topic shifts can also be found when noises or external events interrupt a conversation. We investigate how topics are organised in dialogue and how they relate. Focusing on the topical structure implies to investigate the topic shifts mechanisms. Indeed, while some methods such as topic modelling~\cite{kherwa2019topic} allow for an abstract topic identification, they do not reveal the topical organisation. The way people perceive the current topic is also central to conversation structure as what they consider on or off topic and the places where they would accept a topic shift determine the direction a dialogue can take. Topics hence help to build the structure of the interaction.


Analysing topics in dialogue hence requires conversations long enough to contain several topics and types of topic shifts. The current topic can change abruptly with more or less explicit markers or more gradually. 
Collecting such datasets and annotating topics is challenging~\cite{purver2011topic}. 
Thus for our study we would like to build a dialogue corpus suitable for topical analysis, \textit{i.e.}, where topics would be easier to identify than in entirely casual exchanges, while giving limited constraints to the participants to keep the dialogue as natural as possible. We also want several types of topic shifts to happen. 
Even though oral face-to-face exchange is the most complete form of dialogue, it is also the most complicated to collect due to material and human constraints. Therefore we chose to collect our corpus through a written messaging tool similar to the one developed by \citet{healey2009dialogue}.

In this paper we present the messaging tool we developed for this dialogue collection and outline our experimental plans. We then briefly discuss the pilot study conducted to assess the quality of our tool and finish by drawing conclusions from it.


\section{Method}


    We are interested in how people perceive and negotiate topics in dialogue. Participants carry out a conversational task which allows free conversation, while still being likely to produce several topics  within one domain: the balloon task~\cite{breitholtzReasoningMultipartyDialogue2021}. 
    In this task, passengers are asked to discuss and reach a decision in a moral dilemma  where a balloon with four passengers will crash unless one is sacrificed. Each passenger is valuable for different reasons: the pilot, the pilot's pregnant wife, a child prodigy, and a doctor on the verge of discovering a cure for cancer. 

    This task enables us to identify different sub-topics more easily than casual dialogue. It also allows us to create varying interpretations of the current topic for the participants by switching some task-related words (\textit{e.g.,} ``doctor'' and ``pilot''). We can then see what strategies the participants develop to reach a common understanding. 

    
    Before the data collection, we obtained approval from our university's ethics committee because of the number of participants and the need to share their conversations.
    Meanwhile, we conducted a pilot study to ensure the quality of the tool 
    and to test modifications on the messages (see \cref{sec:pilot}).



\section{The Tool}

Our goal is to manipulate participants' exchanges in written dialogues. To achieve this, we developed a conversational tool using the text-message application Element, which is based on Matrix, a real-time open communication protocol.
%
It is available as a web, desktop, and mobile application which makes it convenient to use for the participants. 
It also allows us to host our own server and keep full control on all of
the experiment data.

\begin{figure}[h]\label{fig:modifs}
    \centering
    \includegraphics[width=\textwidth]{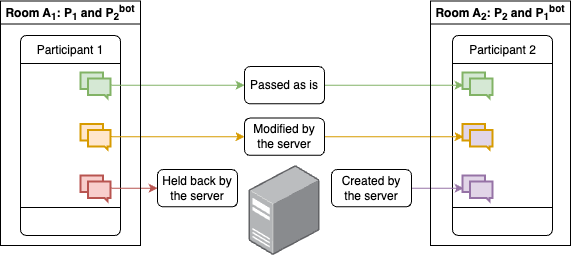}
    \caption{Possible manipulations of the messages with the tool. The \textit{bot} in each room enables the server to be notified of the incoming messages, and to send them back in other rooms.}
\end{figure}

Our tool consists of Python scripts that connect to the server, create accounts and chat rooms for participants, and monitor conversations to modify messages.
Each participant has a real account and an ``associated bot account''.
For each
conversation a participant takes part in, a room is created with them and the bots associated
with the other participants. 
For a two-person conversations, two rooms are created (a real and a bot one).
A message is only received by the other participant's bot. This allows the scripts to process the message and relay it back to the other room as if the bot sent it.
\cref{fig:modifs} illustrates this process. This method enables us to modify the
messages before transmission. Since everything happens in separate rooms,
the sender believes their messages were sent normally and is unaware of the manipulations.
The scripts also store all conversation data in a SQLite database.

Another advantage of this framework is the ability to add widgets to the rooms, \textit{i.e.}, web pages that can be interacted with while chatting. We could use them to create more interactive tasks, for instance, cooperative mini-games.

\section{Pilot study on the Balloon Task}\label{sec:pilot}
    
     We recruited 12 colleagues who participated each in 3 dyadic dialogues (18 pairs). While they were not directly involved in the design of the experiment, some knew their messages might be modified and had a vague idea of the research question. Thus, this pilot does not provide material for topical analysis but confirms the feasibility of our experiment 
    and the effects of the modifications.
    Participants did not know their conversation partners and never had the same one twice. Despite close collaboration in recruitment, 5 pairs did not complete their task, a critical factor for actual data collection.

    We tried several modifications to find the most effective ones. 
    Some modifications related directly to the task, others degraded the utterances independently from the task.

    For the groups with deteriorated sentences, we experimented with removing all verbs, nouns, adjectives, or \textit{stopwords}. These modifications were very disruptive in some cases, such as outputting an almost empty message, but participants quickly developed methods to use the words they wanted, such as changing some letters into digits or separating syllables with spaces. Those changes are thus not suitable for our study since they do not encourage the participants to reach an agreement on the meaning of their sentences. 

    In the groups where task-related words were \emph{removed}, similar behaviours were observed. However, in groups where those words were \emph{switched} for other task-related words, participants often had to discuss what their partner meant. For example, in a group that agreed to save the doctor, a message suggesting sacrificing the pilot was changed to sacrificing the doctor, creating confusion. This type of behaviour is interesting for our study.

    An unexpected discussion occurred in one of the groups where task-related words were switched for others. The participants discussed sacrificing the doctor, and one mentioned it would be acceptable unless the balloon had cancer. This sentence may seem like the result of a modification, but since participants had developed mechanisms to circumvent them already, it confused the receiver even more.



\section{Discussion}
In this paper we presented a dialogue collection experiment that aims at investigating topics and the impact of topic shifts in conversation, and building a corpus that can be used for topical analysis in conversation. Our tool allows for very flexible modifications of the messages sent by the participants. While various changes can be interesting for dialogue analysis, our future data collection will focus on modifications making the representation of the topics different for both speakers and forcing them to explicitly agree on the discussion topic.

We will centre our first experiment on dyadic conversations and hope that we will be able to recruit about 60 French speaking and English speaking participants. We also want to extend this experiment to include Swedish data and multi-party conversations. 
In the future we intend to collect dialogues where participants are encouraged to re-raise past topics or must cover several topics in a short time, allowing us to analyse other mechanisms of topical organisation.

\bibliographystyle{acl_natbib}
\bibliography{anthology,custom}

\onecolumn
\appendix

\section{Appendix: Description of the Conversational Tool}

\end{document}